\documentclass[conference]{IEEEtran}
\IEEEoverridecommandlockouts
\usepackage{cite}
\usepackage{multirow}
\usepackage{amsmath,amssymb,amsfonts}
\usepackage{algorithmic}
\usepackage{graphicx}
\usepackage{textcomp}
\usepackage{xcolor}
\def\BibTeX{{\rm B\kern-.05em{\sc i\kern-.025em b}\kern-.08em
    T\kern-.1667em\lower.7ex\hbox{E}\kern-.125emX}}
\begin{document}

\title{An Approach for Combining Multimodal Fusion and Neural Architecture Search Applied to Knowledge Tracing}
\author{\IEEEauthorblockN{1\textsuperscript{st} Xinyi Ding}
\IEEEauthorblockA{\textit{School of Computer and Information Engineering} \\
\textit{Zhejiang Gongshang University}\\
Hangzhou, China \\
xding@zjgsu.edu.cn}
\and
\IEEEauthorblockN{2\textsuperscript{nd} Tao Han}
\IEEEauthorblockA{\textit{School of Computer and Information Engineering} \\
\textit{Zhejiang Gongshang University}\\
Hangzhou, China \\
hantao@zjgsu.edu.cn}
\and
\IEEEauthorblockN{3\textsuperscript{rd} Yili Fang}
\IEEEauthorblockA{\textit{School of Computer and Information Engineering} \\
\textit{Zhejiang Gongshang University}\\
Hangzhou, China \\
fangyl@zjgsu.edu.cn}
\and
\IEEEauthorblockN{4\textsuperscript{th} Eric Larson}
\IEEEauthorblockA{\textit{Lyle School of Engineering} \\
\textit{Southern Methodist University}\\
Dallas, USA \\
eclarson@lyle.smu.edu}
}

\maketitle

\begin{abstract}
Knowledge Tracing is the process of tracking mastery level of different skills of students for a given learning domain. It is one of the key components for building adaptive learning systems and has been investigated for decades. In parallel with the success of deep neural networks in other fields, we have seen researchers take similar approaches in the learning science community. However, most existing deep learning based knowledge tracing models either: (1) only use the correct/incorrect response (ignoring useful information from other modalities) or (2) design their network architectures through domain expertise via trial and error. In this paper, we propose a sequential model based optimization approach that combines multimodal fusion and neural architecture  search within one framework. The commonly used neural architecture search technique could be considered as a special case of our proposed approach when there is only one modality involved. We further propose to use a new metric called time-weighted Area Under the Curve (weighted AUC) to measure how a sequence model performs with time. We evaluate our methods on two public real datasets showing the discovered model is able to achieve superior performance. Unlike most existing works, we conduct McNemar’s test on the model predictions and the results are statistically significant. 
\end{abstract}


\section{Introduction}

For a given educational domain, the learning goal for a student is to master a set of skills (knowledge components) that are usually designed by experts or discovered automatically by computational models \cite{piech2015deep}.  For instance, in Geometry, calculating the area of a circle is considered one skill in a larger learning goal. Based on the student's proficiency with varying skills, customized learning materials could be provided, potentially improving learning efficiency. The process of tracking the mastery level of different skills is called Knowledge Tracing (KT), and it is a key component for building adaptive learning systems. During the interaction between one student and an adaptive learning system, data like the sequence of responses to some questions, the time one spent on each specific question, hints that asked, etc. will be logged. All these interaction data between students and the adaptive learning system can be used to build KT models---however, the most often used data is a single sequence of correct/incorrect responses, as indicated by previously proposed knowledge tracing models\cite{corbett1994knowledge, pavlik2009performance, piech2015deep, zhang2017dynamic, pandey2019self, ding2020automatic}. In these works, the objective of the model is to analyze a sequence of one student's responses and predict which future questions the student will answer correct or incorrect. 

Various methods for KT have been proposed. Bayesian Knowledge Tracing (BKT) \cite{corbett1994knowledge} is a type of Hidden Markov Model (HMM), with latent variables modeling the skills and observed variables modeling the student responses. Performance Factors Analysis (PFA) \cite{pavlik2009performance}  calculates the accumulated learning as a function of the number of success attempts and failure attempts for different skills. 
%
The empirical success of deep learning in other fields motivated researchers in the learning science community to incorporate deep neural networks for knowledge tracing. Deep Knowledge Tracing (DKT) \cite{piech2015deep} typically uses a recurrent neural network which takes one-hot encoding of the skill tag and correctness as input. The output vector corresponds to the mastery level of different skills. Dynamic Key-Value Memory Network \cite{zhang2017dynamic} uses one static key memory to encode the skill tags and one value memory to store the mastery level of skills. The value memory is updated accordingly after every attempt. Other works like Paney \textit{et al.} incorporates attention mechanisms for knowledge tracing \cite{pandey2019self}. Ding \textit{et al.} exploits neural architecture search for recurrent cell discovery for knowledge tracing \cite{ding2020automatic}. 

When building Knowledge Tracing models, all above mentioned works use only the correct/incorrect response history of a student. However, most existing adaptive learning systems or intelligent tutoring systems usually provide additional information about the interactions of a student. Along with the correct/incorrect responses, these systems monitor information like the time spent on each item, whether one student asks for a hint or not, accommodations used, etc. \cite{feng2006addressing, stamper2010kdd, koedinger2010data}.
Some existing works have shown that incorporating more features (we use features and modalities interchangeably in this paper) have the potential of improving model performance \cite{ding2019eduaware, zhang2017incorporating, yang2018implicit}. For example, Zhang \textit{et al.} \cite{zhang2017incorporating} concatenated correct/incorrect response together with additional discrete features into a vector and used Auto-Encoding to  learn a representation with lower dimensionality. In parallel, Yang \textit{et al.} \cite{yang2018implicit} proposed a similar work. However, instead of discretizing features first, they used a decision tree which could handle both continuous and discrete input. The decision tree will first take in all the features and then output a prediction about the next response.

Most existing deep neural network based knowledge tracing models only use one modality and manually crafted architectures, with some exceptions either using one modality and neural architecture search\cite{ding2020automatic}, or using multimodal inputs with manually designed architectures \cite{zhang2017incorporating, yang2018implicit}. In this work, we try to bridge the gap by proposing an approach that combines multimodal fusion and neural architecture search within one framework. Our contributions are summarized as follows:

\begin{itemize}
    \item We propose an approach that combines multimodal fusion and neural architecture search within one framework. The commonly used neural architecture search technique could be considered as a special case of our proposed approach when there is only one modality involved. 
    \item We propose to use a new metric called time-weighted area under the curve (wAUC) that could measure how one model performs with time.
    \item We evaluate our method on two public real datasets. Our method is able to find neural architectures having superior performance. We run McNemar's test on the predictions and the results are significantly different. 
\end{itemize}

\section{Related Work}
Knowledge Tracing is a well established research area with roots in traditional modeling like those based on item response theory models \cite{gonzalez2014general, khajah2014integratinglatent}, Bayesian Knowledge Tracing (BKT) \cite{corbett1994knowledge}, and Performance Factors Analysis (PFA) \cite{pavlik2009performance}, and has evolved to more recent deep neural network based models \cite{piech2015deep, zhang2017dynamic, pandey2019self, ding2020automatic}. In this section, we focus on the most recent works based on deep neural networks. Compared to traditional statistical models, these works have shown state-of-the-art performance in predicting student responses. 

\subsection{Knowledge tracing based on deep neural networks}

The first work to propose Deep Knowledge Tracing (DKT) was from Piech \textit{et al.} \cite{piech2015deep}. The DKT model uses a standard recurrent neural network architecture, for example LSTM\cite{hochreiter1997long}. Each student response that is analyzed by the model contains information about what skill the question is designed to assess (the skill ID). The input to this model is a one hot encoding of the skill ID crossed with the correctness of the problem. For example, if there are $M$ skills, the size of the input vector will be $2*M$. The authors tried other options, like encoding the skill ID and correctness separately, but results proved unsuccessful. The output layer of the model is a vector of probabilities predicting if the student could answer each of $M$ skills correctly.  However, this output vector is not trained as a whole. For each student, the DKT model takes the combination of the skill ID and correctness from the previous problem to predict the next problem. The next problem ID will be used to choose the corresponding element in the output vector, based on which the cost function is created. Also, the DKT model does not distinguish different students. 

Dynamic key-value memory network for knowledge tracing (DKVMN) \cite{zhang2017dynamic} is based on the work from Weston \textit{et al.} \cite{weston2014memory}.  DKVMN has one static key memory $M^k$, which is the embeddings of all skills. The content of $M^k$ does not change with time.  DKVMN also has one dynamic value memory $M_t^v$ for storing the current mastery level of corresponding skills. The content of $M_t^v$ is updated after each response. There are two stages involved in the DKVMN model. In the read stage, a query skill $q_t$ is first embedded to get $k_t$, then a correlation weight and the mastery level of skill $q$  is calculated using:
\begin{equation*}
     w_t(i) = \text{softmax}(k_t^TM^k(i))
\end{equation*}
\begin{equation*}
    \textbf{r}_t = \sum w_t(i)M_t^v(i)
\end{equation*}
 The authors concatenate the query skill $q_t$ with $\textbf{r}_t$ to get the final output $p_t$ arguing that the difficult level of each skill might be different. The second stage is to update the memory network $M_t^v$. The embedding of the combination of the skill query $q_t$ and the actual correctness $r_t$ is used to create an erase vector $\textbf{e}_t$ and an add vector $\textbf{a}_t$. The new value matrix is updated using:
 \begin{equation*}
      \Tilde{M}_t^v(i) = M_{t-1}^v(i)[\textbf{1} - w_t(i)\textbf{e}_t]
 \end{equation*}
\begin{equation*}
    M_t^v(i) = \Tilde{M}_t^v(i) + w_t(i)\textbf{a}_t
\end{equation*}
Pandey \textit{et al.} \cite{pandey2019self} proposed using self attention mechanism for solving the sparsity issues recurrent neural networks could face, since the next prediction only depends on a few most relevant attempts in the past.  Most existing deep neural network architectures for knowledge tracing are designed manually by trail and error or from intuition. Ding \textit{et al.} \cite{ding2020automatic} proposed using reinforcement learning for the automatic design of recurrent cells for knowledge tracing. In their approach, a recurrent cell is encoded as outputs of an agent which is another recurrent neural network. The performance of the generated models are treated as rewards that could guide the agent to output better models. In our work, we also use neural network search to automatically look for the best architectures---however, unlike the above mentioned works, we incorporate this search with additional feature modalities.

\subsection{Multimodal deep knowledge tracing}

Most online intelligent tutoring platforms track not only  the correctness of a response, but also quantities such as how many attempts one student has tried, what is the result of the first attempt, how much time was spent on an item, and ``does one student ask for a hint?'', among many other factors.    

Based on the DKT model, Zhang \textit{et al.} \cite{zhang2017incorporating} tried including more features like \textit{response time}, \textit{attempt count}, \textit{first action}, and others. The only difference between their model and the DKT model is the input. Features are first discretized if they are continuous, then these features are concatenated to form one input vector. However, this input vector grows exponentially with the number of features. Thus, they used neural network Auto-Encoding \cite{hinton2006reducing} to learn a representation with lower dimensionality.  They evaluated this model on different datasets and the results show improved performance. However, using simply concatenation to incorporate these features has a number of downsides including increased memory footprint and exponential growth in the number of parameters needed for training the encoding. 

Yang \textit{et al.} \cite{yang2018implicit} proposed a similar work. Instead of discretizing features first, they used a decision tree which could handle both continuous and discrete inputs. The decision tree first analyzes all the features and then outputs a prediction about the next response. The predicted response is combined with the actual response to form a 4 bit one-hot encoding (1000, 0010, 0100, 0001 representing true positive, false positive, false negative and true negative respectively). This one hot encoding then is concatenated with the original one hot encoding of correctness and exercise tag to form the final input. They evaluated this technique on several datasets and the results also show improved performance compared with using only the correctness as input. 

However, the way these models combine different modalities neglects the hierarchical and representational learning ability of deep neural networks. Thus, our work builds upon the evidence from \cite{zhang2017incorporating,yang2018implicit} that incorporating modalities can increase performance, but we further incorporate methods to find the correct fusion of these modalities, rather than relying solely on ad-hoc or expert designs.

\section{Methods}

\subsection{Neural architecture search}
\label{sec::nas}

\begin{figure}[t]
\centering\includegraphics[width=0.6\linewidth]{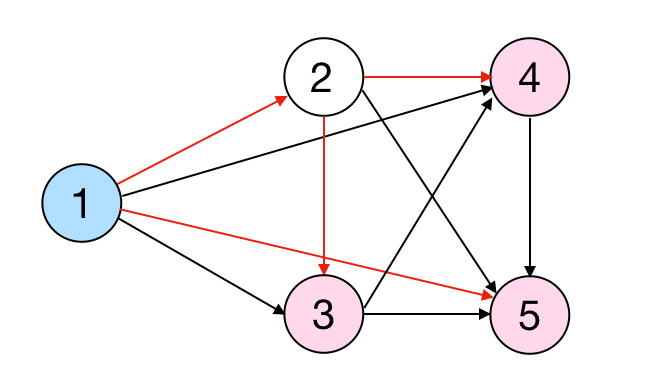}
\caption{Search space for a recurrent cell,  higher level nodes (layers) are fully connected with lower level nodes. Red arrows  indicate a sampled architecture. Parameters are shared among all sub models.}
\label{fig::param_share}
\end{figure}

A lot of successful deep neural network architectures are developed by experienced experts through trial and errors. Apparently, this process is tedious and not very efficient. Neural Architecture Search (NAS) \cite{pham2018efficient, zoph2016neural, liu2018progressive} aims to find good architectures for some specific task automatically and has gained more and more attention in the past few years. It has been shown that some architectures found through NAS have achieved better performance than those developed manually by human experts \cite{elsken2019neural}.  Evolutionary algorithm\cite{back1996evolutionary} is one of those earliest used for NAS. The idea is to represent the neural network architecture with a fixed size vector. For example, the first element in this vector could be the type of the first layer (conv layer, max pooling layer, etc). Off springs are generated using mutations of the parents vectors. The generated architectures are evaluated based on some metrics (accuracy on a validation set). This process keeps going until we have found some good architectures or we hit the maximum number of iterations. If we look more closely, we could further decompose the model finding problem into architecture search and parameters optimization. In such cases, Evolutionary algorithms are usually used together with gradient based methods, in which Evolutionary algorithms are used for finding the architecture, and gradient based methods are used to optimize the weights. Other techniques used for NAS include Reinforcement Learning (RL) \cite{zoph2016neural, pasunuru2019continual, pham2018efficient},  gradient based methods\cite{wang2020apq, saikia2019autodispnet, dong2019one}, sequential model based optimization (SMBO) \cite{liu2018progressive}, etc. For a comprehensive survey of NAS, readers could refer to this survey \cite{elsken2019neural}.

\begin{figure*}[t]
\centering
\includegraphics[width=0.9\linewidth]{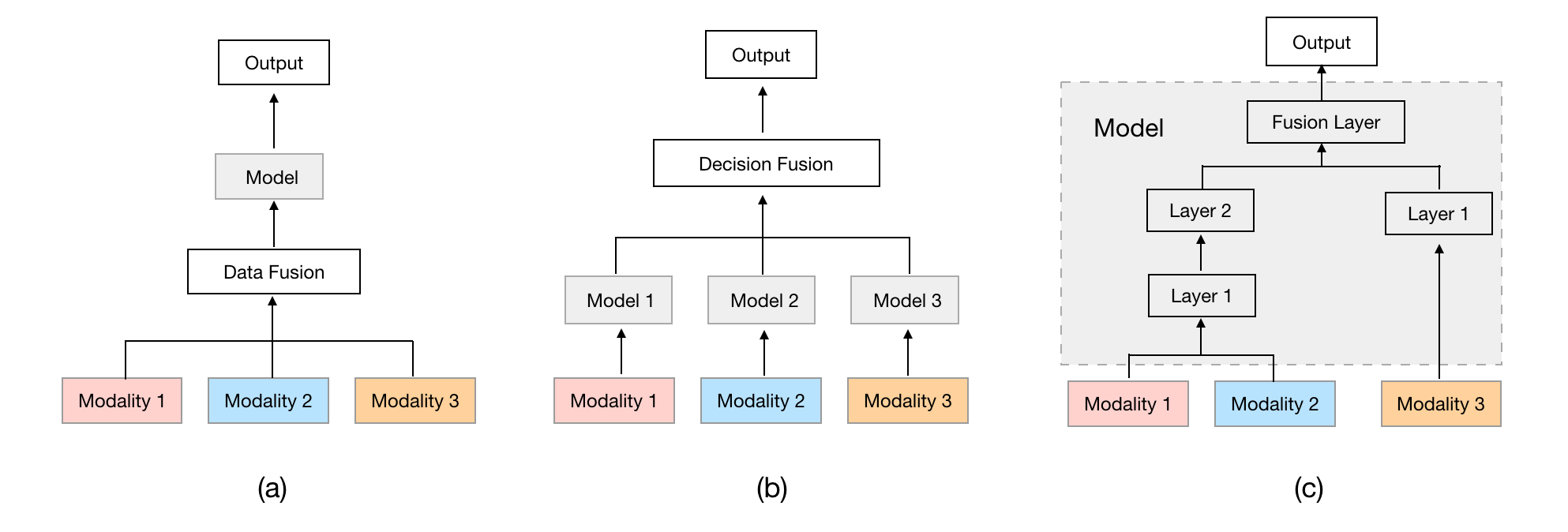}
\caption{Different levels of fusion. (a) Early level fusion, (b) Late level fusion, (c) Intermediate level fusion, as categorized by \cite{ramachandram2017deep}.}
\label{fig::mml}
\end{figure*}

Neural architecture search could be computationally demanding if used directly, thus several mitigations have been proposed. Parameters sharing \cite{pham2018efficient} is a technique to share parameters among generated models. The idea is not to reinitialize the parameters for each new generated model, instead, subsequent models use the parameters from previous trained models. Some works\cite{zoph2016neural} take a greedy approach, looking for one unit that could be stacked to form the final architecture. For example, for a recurrent architecture, we could search for the best recurrent cell type or best activation function to use in the cell. Sequential Model Based Optimization (SMBO) \cite{liu2018progressive} gradually unfold more complex networks and use a surrogate function to predict the performance of potential models. Only these models with predicted high performance will be trained, thus reduce search space.

In this work, we look for the best recurrent cell that could be further stacked into our model. We regard cell architecture search as the process of sampling a sub-graph from the global graph as shown in Figure \ref{fig::param_share}. This idea is similar to the one in \cite{pham2018efficient}. Each node (layer) is fully connected to its previous nodes. The outputs of earlier nodes will be used as inputs to later nodes and a later node could choose the output of any its previous node.  A sampled sub graph (model) is indicated as red arrows in this figure. Node 1 is the input node, node 3, node 4, and node 5 are leaf nodes and their outputs will be combined to generate the final predictions. A fully connected layer involves the operation of a matrix multiplication and activation function. For the input node, the following computations are performed:
\begin{equation}
    c_1^{t} = \phi(x^t\cdot W_0^{(x,c)} + h_0^{t-1} \cdot W_0^c)
\end{equation}
\begin{equation}
        h_1^t = c_1^t*f_1(x^t\cdot W_0^{(x, h)} + h_0^{t-1} \cdot W_0^h) + (1 - c_1^t)*h_0^{t-1}
\end{equation}
where $\phi$ and $f_1$ are both activation functions. $\cdot$ refers to dot products, $*$ refers to element wise vector multiplication.  For subsequent layers, the following computations are performed: 
\begin{equation}
    c_l^t = \phi(h_{j,l}^t \cdot W_{j,l}^c)
\end{equation}
\begin{equation}
    h_l^t = c_l^t*f_l(h_{j,l}^t \cdot W_{j,l}^h) + (1 - c_l^t)*h_{j,l}^t
\end{equation}
where $\phi$ and $f_l$ are activation functions, we use $sigmoid$ function for $\phi$, and $W_{j,l}$ are the feedforward weights from node $j$ to node $l$. 
We get the final output by averaging the leaf nodes. In the case of the sampled architecture shown in Figure \ref{fig::param_share}, we get:

\begin{equation}
    h_o^t = (h_3 + h_4 + h_5) / 3
\end{equation}

To save search time, in this study we use the parameters sharing technique \cite{pham2018efficient}.  The idea of parameters sharing is that weights are only initialized once at the very beginning of search and are shared among all sub-models. Thus, for each new generated model, we do not reinitialize the weights, but start from the last training period. We also use Sequential model based optimization (SMBO) \cite{liu2018progressive} technique to generate new models. SMBO assumes that the search space could be gradually unfolded from simple to complex. When constructing a recurrent cell, the architecture becomes more and more complex when more layers are added. Lying in the heart of SMBO is a function that could predict the accuracy of a generated model. This function is usually called the ``surrogate function.'' The search starts from the simplest case (for example only one layer), after all possible models are trained. The architectures and their accuracies will be used to train the surrogate model. Then more complex models are unfolded (by adding more layers). When the search space becomes too large to train all possible models, a sampling process is taken. In this case, the surrogate function will be used to predict the accuracies of these models. Then $k$ models are sampled (with higher accuracy, higher the probability to be sampled, allowing exploration of numerous architectures). Then the sampled $k$ models are trained. Their accuracies and architectures are then used again to update the surrogate function. Thus, it is an alternative process. Compared with reinforcement learning, SMBO has a simpler implementation and has increased stability.  Because the sampled architectures could be encoded as a sequence of list like $[[p_1, a_1], [p_2, a_2], ...[p_i, a_i]]$, where $p_i$ stands for previous node and $a_i$ stands for activation function,  we could use a recurrent neural network as our surrogate function.

\subsection{Multimodal fusion}

When considering the use of multimodal inputs, the key insight is about crafting a strategy for fusion of each mode. Figure \ref{fig::mml} shows three possible ways of fusing different modalities. Prior to the appearance of Deep Learning, there are mainly two ways that multi modalities could be used in a machine learning model. The first one is to combine different modalities at the very beginning (Figure \ref{fig::mml} (a)). For example, simply concatenating different modalities into large feature vector. The hope was that the model could learn all useful information from this vector. This method is also called early fusion. Another way is to train a different model for each modality, then combine the outputs from these models to make a final decision (Figure \ref{fig::mml} (b)). This is similar to the ensemble modeling and also known as late fusion or decision level fusion. Neither early fusion nor late fusion is proved to perform better than the other in all situations. Which methodology to use highly depends on the application. 

One of the reasons why deep learning is so successful on perceptual tasks is its ability to learn hierarchical representations. Features are learned from data automatically instead of manually crafted.  Figure \ref{fig::mml} (c) shows an example of using deep neural networks for intermediate fusion. As we can see intermediate fusion allows features from different modalities to be gradually integrated. Empirical results have shown the advantages of intermediate fusion. However, to use intermediate fusion, we still need to decide the fusion architecture first (which modalities to fuse at which level).  Such kind of intermediate fusion using deep learning is also called deep multimodal learning \cite{ramachandram2017deep}. In this study, instead of manually designing the fusion architecture for deep neural networks, we use neural architecture search to automatically look for the best fusion structure.

\subsection{Combining neural architecture search and multimodal fusion}

In this section, we discuss different ways of combining neural architecture search and multimodal fusion within one framework. We will first consider the simplest case which is to search for the best representation for one skill. 

\subsubsection{Look for the best representation for one skill}

\begin{figure}[t]
\centering
\includegraphics[width=0.7\linewidth]{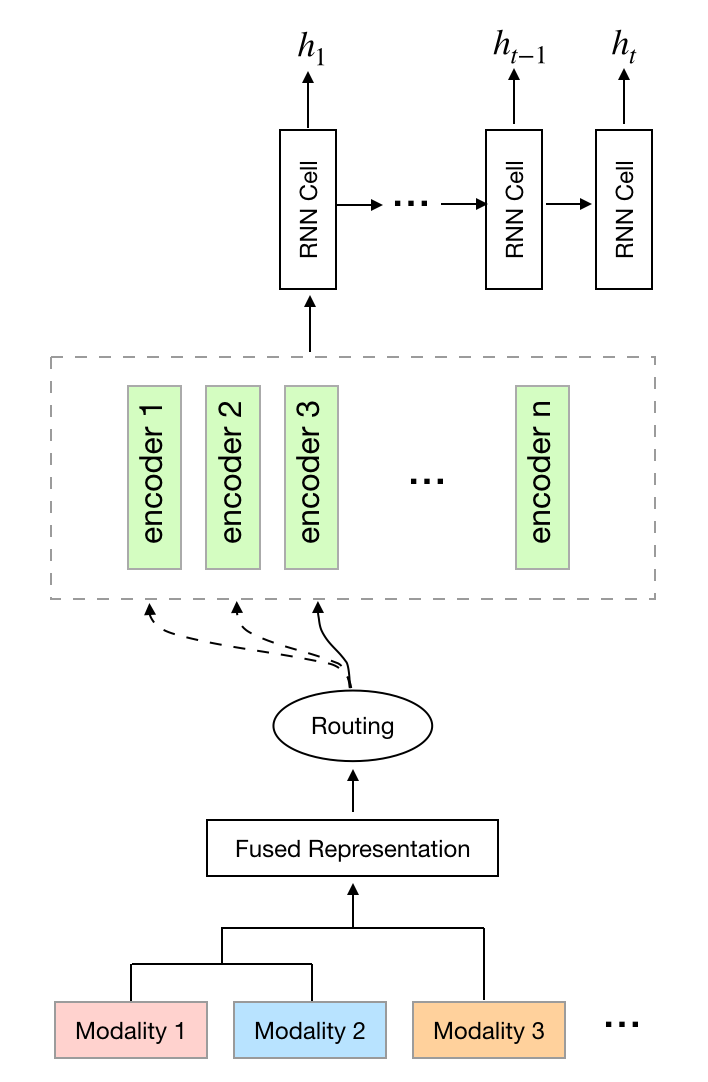}
\caption{NAS for multimodal fusion with each skill has its own separate encoder. The representation from the encoder goes through a regular recurrent neural network (DKT model). In this case, the modality 1 and modality 2  will be fused first, then modality 3 (the activation functions used are not shown in this figure). }
\label{fig::search_1}
\end{figure}

A simple, baseline approach is to first learn a representation that contains information from all modalities and then send this representation to the recurrent neural network (DKT model). Our first proposed architecture is shown in Figure \ref{fig::search_1}. Two significant differences from the DKT is that we use multimodality and we have a separate encoder (fully connected layer) for each skill. Using a separate encoder for each skill also means we do not consider skill ID as one modality, like others. Piech \textit{et al.} \cite{piech2015deep} also discussed to encode the skill ID and correctness separately. However, they did not achieve results as performant as using one-hot encoding. Also, they did not give the details about how they conducted separate encoding.  Using a separate encoder has at least two benefits here. Firstly, we argue that each skill is different from others (to some degree), thus using a separate encoder allows custom transformation of a skill. Secondly, it is more convenient to combine more modalities. The final fused representation could be processed through the corresponding encoder and without needing to attach the skill ID to each modality. To make sure the performance improvement is due to the actual fusion structure, not because of the using of separate encoders, we did an experiment only using the response and separate encoders and find the results are similar to the ones from the DKT model. 

Each modality will be first embedded using an embedding layer. We use embedding size 100, this is the same for all modalities. We consider the problem of which modalities to fuse at which order and what activation functions to use. The architecture could be encoded as a list of sequence of numbers. For example,  the fusion architecture [[$m_1$, $m_2$, $f_0$], [$m_4$, $m_3$, $f_1$]] could be interpreted as: Fuse modality $m_1$ and modality $m_2$ first using activation function $f_0$ (e.g., hyperbolic tangent). Then fuse in another modality $m_3$ (here $m_4$ means the fused representation of modality $m_1$ and modality $m_2$) using the activation function $f_1$. From this perspective, the fusion architecture search process also functions as a features selection process. Thus, harmful or noisy features will be automatically filtered out. In this work the number of modalities and types of activation functions used is somewhat small---however, the proposed methodology could be easily extended to any number of modalities and activation functions. Adding more modalities could be considered as gradually increasing the complexity of the sampled space. 

\subsubsection{Extend sub-graph sampling process to multimodality}

\begin{figure}[t]
\centering
\includegraphics[width=0.6\linewidth]{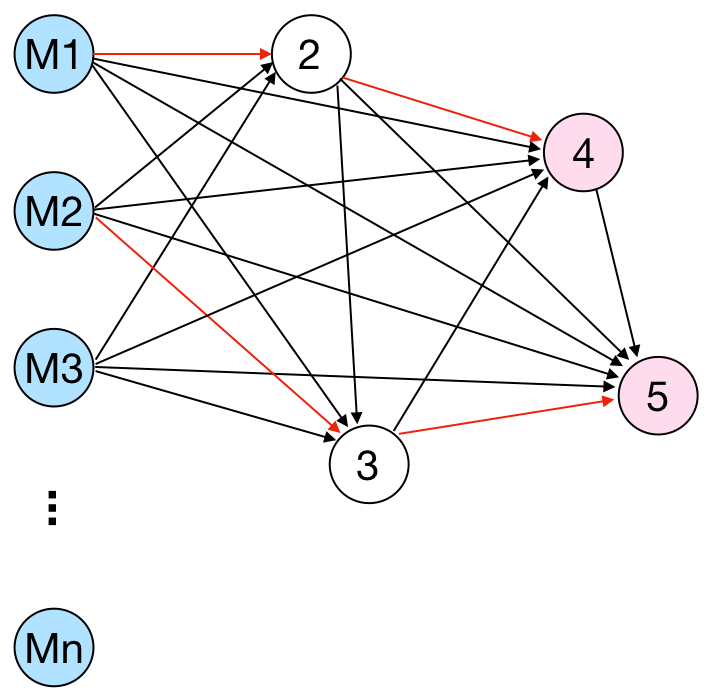}
\caption{Extend sub-graph sampling to multimodality. Red arrows indicate a sampled architecture using two modalities.  Modality $m_1$ is used as input to node 2 and then further node 4. Modality $m_2$ is used as input by node 3, then further by node 5. Node 4 and node 5 are leaf nodes here and their outputs will be combined as the final output.}
\label{fig::search_2}
\end{figure}

In the previous section, we applied NAS for multimodal fusion search while keeping the cell architecture fixed (LSTM). In this section, we take one step further and consider the possibility of combining multimodal fusion search with cell architecture search. We accomplish this by extending the sub-graph sampling approach discussed in section \ref{sec::nas}. There are few works that combine multimodal fusion search and architecture search within one methodology. One exception is the work from P{\'e}rez-R{\'u}a \textit{et al.}\cite{perez2019mfas}. However, they only considered the case of two modalities. They used two pre-trained networks for two different modalities. Each modality is represented by a pre-trained multi-layer neural network. Their goal is to extract representations from different layers from these two modalities and fuse them in a specific order to achieve good prediction performance.  Strictly speaking, their work is representation search from two pre-trained networks, not architecture search. 

Now, we focus on how can we extend the sub-graph sampling process of \cite{pham2018efficient} to multimodality search. Our proposed architecture is shown in Figure \ref{fig::search_2}. Similarly, each node (layer) is fully connected with its previous nodes and it could choose the output from any of these previous nodes. The difference is, instead of one input node, we have multiple input nodes with one for each modality. Again, we first apply an embedding layer for each modality to achieve a representation easier for further processing. Each skill has its own encoder, one for each modality. We can see this search space is a superset of the one previously discussed. Within this methodology, any amount of modalities could be fused at any layer. The sampled architecture could also be represented by a list of sequence of numbers [[$p_1$, $a_1$], [$p_2$, $a_2$],...[$p_i$, $a_i$]], where $p_i$ refers to the previous node and $a_i$ refers to the activation function. The red arrows in Figure \ref{fig::search_2} indicates a sampled architecture. Modality $m_1$ is used as input to node 2 and then  further node 4. Modality $m_2$ is used as input by node 3, then further by node 5. Node 4 and node 5 are leaf nodes here and their outputs will be combined as the final output.

\subsection{A new metric for Knowledge Tracing}
Area under the curve (AUC) and coefficient of determination ($r^2$) are two standard metrics used for knowledge tracing. However, neither metric is designed to be used to measure how one model performs over a sequence. That is, does the model perform better after it sees more responses from one student? Thus, we propose to use weighted AUC (wAUC) for this purpose \cite{li2010weighted}. For a series of items to be predicted $i \in \{1, ..., n\}$. There is a series of corresponding weights $\{w_1, w_2, ... w_n\}$. Let $S_1 = \{i: y_i = 1\}$ be the set of positive examples. $S_0 = \{i: y_i = 1\}$ be the set of negative examples. Weighted false positive rate and weighted true positive rate could be defined as follows:

\begin{equation*}
    FPR = \frac{1}{W_0}\sum_{i\in S_0} \textit{I} [y_i = 1]w_i
\end{equation*}
    
\begin{equation*}
    TPR = \frac{1}{W_1}\sum_{i\in S_1} \textit{I} [y_i = 1]w_i
\end{equation*}

\noindent where, $W_0 = \sum_{i \in S_0} w_i$ is total negative weights and  $W_1 = \sum_{i \in S_1} w_i$ is the total positive weights. Thus, the weighted ROC curve could be plotted for all thresholds. Weighted AUC could be used in cases in which we want to emphasize low false positive rate, etc.  In our case, we employ it to measure how the knowledge tracing model performs with emphasis on time. We choose to assign more weights to the most recent responses. That is, we are not concerned that when the model first begins predicting student responses. In this work, we use a simple strategy that is to increase the weight by one each time the student sees the same skill. This strategy of weight assigning is simple, but sensitive to measures that occur later in the sequence. We believe more mature weights assigning strategies using domain knowledge exist. But, as an initial analysis, this simple metric and weight assigning strategy allows us to measure how one model performs with time. We hope this work could inspire more knowledge tracing model measurement research in the future. 

\section{Experiments}

\begin{figure*}[t]
\centering
\includegraphics[width=0.9\linewidth]{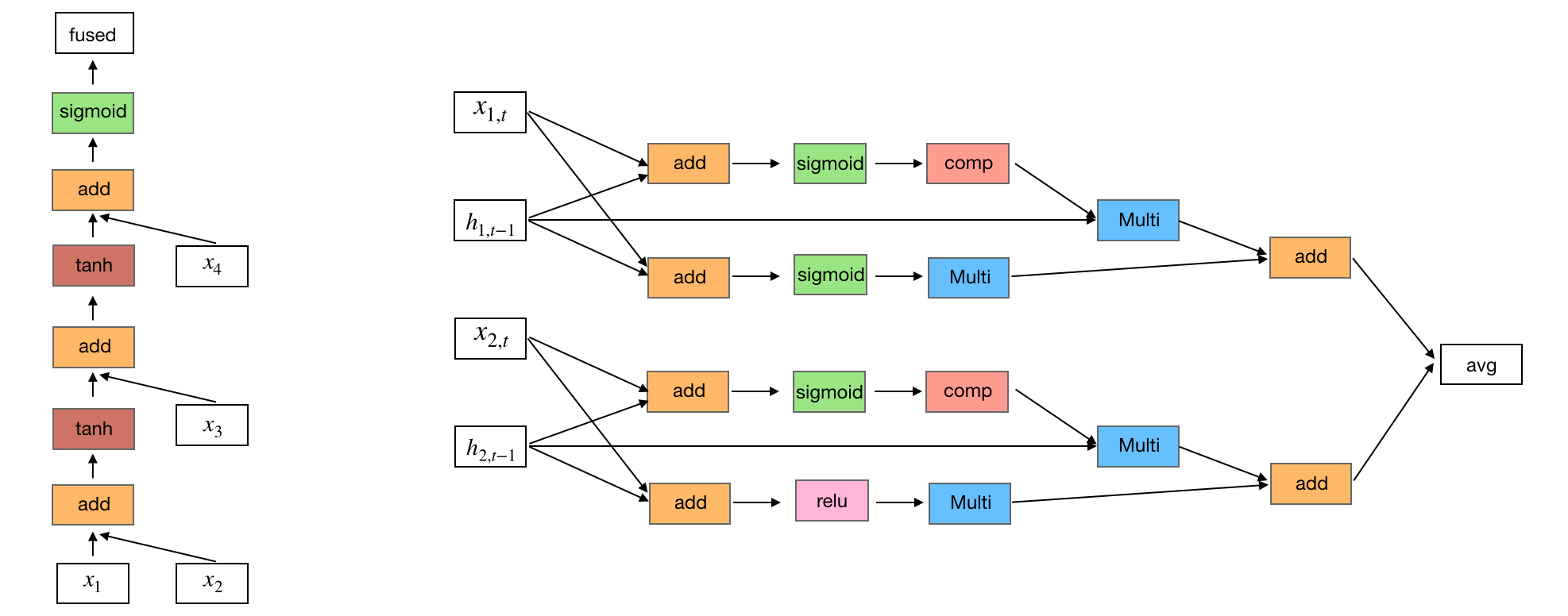}
\caption{The discovered best architectures for knowledge tracing. \textbf{Left:} Multimodal fusion search (FS) and use the fixed LSTM recurrent cell (Fig.\ref{fig::search_1}). \textbf{Right:} Extend the sub-graph sampling to multimodality (Fig.\ref{fig::search_2}). Here $add$ stands for element wise addition. $tanh$ is the hyperbolic tangent function, $sigmoid$ is the logistic function, $relu$ is the rectified linear activation function, $Multi$ is the dot production and $comp$ indicates (1 - c) operation. }
\label{fig::archs}
\end{figure*}

We evaluate our proposed approach on two public real datasets. The overall dataset statistics are given in Table \ref{tab::nas_statitics_2}.

\begin{table}[h]
\centering
\caption{Dataset Statistics}
\label{tab::nas_statitics_2}
\begin{tabular}{cccc} 
\hline
                                                                            & Records & Students & Skills  \\ 
\hline
Assistment 09-10                                                            & 337,236 & 3,884    & 123     \\
\begin{tabular}[c]{@{}c@{}}Oli Engineering Statics\\ 2012 Fall\end{tabular} & 344,403 & 566      & 1251    \\
\hline
\end{tabular}
\end{table}

\textbf{Assistment skill builder 09-10}
The ASSISTment system is an online tutoring system originally built on $8^{th}$ grade MCAS test items (mathematics)\cite{feng2006addressing}. The $8^{th}$ grade mathematics test includes the following five domains: The number system, Expressions and equations, Functions, Geometry, Statistics and probability.

\textbf{OLI Engineering statics 2012 Fall:} The Carnegie Mellon's Open Learning Initiative (OLI) is an online platform that provides customized learning\footnote{https://oli.cmu.edu/}. OLI aims for higher education. The OLI Engineering Statics course is the study of methods for quantifying the forces between bodies. The data were collected for the term 2012 Fall.

Both these two datasets contain other modalities than responses. Great care need to be taken when combining these different modalities for knowledge tracing. After removing those modalities that have too many empty values, the ones considered in this study include the following:

\begin{itemize}
    \item \textit{Response}. This is a binary variable indicating either one correct or incorrect answer for the current problem. For models like DKT and DKVMN, this modality is the only one used. 
    \item \textit{Time spent}. The total time spent on the current problem. This is a continuous variable.
    \item \textit{Attempts}. This feature indicates how many opportunities the current student has had for applying the associated skill.
    \item \textit{Hints}. How many possible hints are there for this problem. This one is problem specific.
    \item \textit{First Action}. This is a binary variable indicating whether one student tries to attempt or ask for hint. If this student asked for a hint, the response is set to be incorrect.
\end{itemize}

Note, we do not consider the skill id as a separate modality like other work, instead we use different encoders (fully connected layers) for different skills. 

It is difficult to perform a fair comparison between two different deep neural network models for a number of reasons. Firstly, we only have a limited understanding of how neural networks function or make exact use of training data. Secondly, techniques like batch normalization\cite{ioffe2015batch}, weights initialization, learning rate, regularization, selection of optimizer, and many others can have a significant impact on the performance of models. Can one model with a novel weight initialization technique be considered a different model? What exacerbated this problem is that one set of hyperparameters that works well on one dataset might not work well on another dataset. To explore all possible hyperparameters combination is also impractical. A recent paper \cite{brendel2019approximating} states that a simple model with careful picked settings could achieve better results than many deep neural network models. Thus, the success of complicated models have might due to the fine tuning of hyperparameters, not because they have better decision strategies. In this work, to make a fair comparison, we follow the conventions from other researchers. We did a reasonable amount of hyperparameters tuning (limited only by time and available computation) for the baseline models used in this study and we limit the tuning parameters to learning rate, weight decay, the epilson value of adam optimizer. However, we would not claim that we are using the best hyperparameters combination and we believe it is always possible to improve these numbers shown in Table \ref{tab::all_results} by tweaking a little bit more on these hyperparameters. Thus, we hope readers will focus more on the architecture innovations and other contributions. Despite the fact that there is no significant test on most deep neural network papers, We did McNemar's test on the predictions \cite{McNemar1947}. McNemar's test focuses on the distributions of predictions. The test is applied to a 2x2 table as shown below: 

\begin{table}[h]
\centering
\refstepcounter{table}
\label{tab::mcnemar}
\begin{tabular}{cccc} 
\hline
             & M2 positive & M2 negative & Row total  \\ 
\hline
M1 positive  & a           & b           & a+b        \\
M1 negative  & c           & d           & c+d        \\
Column total & a+c         & b+d         & n          \\
\hline
\end{tabular}
\end{table}

\begin{table*}
\centering
\caption{Comparision of different models for knowledge tracing. SM stands for Single Modality. MM stands for multimodality. SC stands for simple concatenation. FS stands for fusion search. }
\label{tab::all_results}
\begin{tabular}{cccccccc} 
\hline
\multicolumn{2}{c}{\multirow{2}{*}{}}    & \multicolumn{3}{c}{Assistment 09-10}                & \multicolumn{3}{c}{Oli Statics 2012}                 \\ 
\cline{3-8}
\multicolumn{2}{c}{}                     & r2              & AUC             & wAUC            & r2              & AUC             & wAUC             \\ 
\hline
\multirow{3}{*}{SM} & DKT    & 0.1628          & 0.7326          & 0.7367          & 0.4106          & 0.8819          & 0.8622           \\
                    & DKVMN  & 0.1507          & 0.7299          & 0.7354          & 0.3557          & 0.8793          & 0.8614           \\
                    & NAS cell           & 0.1678          & 0.7364          & 0.7408          & 0.4169          & 0.8844          & 0.8661           \\ 
\hline
\multirow{3}{*}{MM} & DKT + SC           & 0.1743          & 0.7371          & 0.7441          & 0.4316          & 0.8884          & 0.8734           \\
                    & DKT + FS           & \textbf{0.1844} & 0.7454          & 0.7493          & 0.4239          & 0.8863          & 0.8651           \\
                    & NAS Extend         & 0.1829          & \textbf{0.7458} & \textbf{0.7545} & \textbf{0.4348} & \textbf{0.8902} & \textbf{0.8779}  \\
\hline
\end{tabular}
\end{table*}

The null hypothesis and alternative hypothesis are:
\begin{equation*}
    H_0: p_b = p_c
\end{equation*}
\begin{equation*}
    H_1: p_b \ne p_c
\end{equation*}

The McNemar's test statistic is:

\begin{equation*}
    \chi^2 = \frac{(b-c)^2}{b+c}
\end{equation*}
$\chi$ has a chi-squared distribution. If the results are significant, it will reject the null hypothesis, in favor of the alternative hypothesis. Thus, the predictions from these two models are considered different. Although these significance testing tools have their own limitations\cite{dietterich1998approximate}. And some of the underlying assumptions might be violated. For example, the dataset we used is not i.i.d data (independently and identically distributed data). They are more like time series data. 

Since the process of neural architecture search is highly time consuming and we have two different datasets. The methodology we took is we use the assistment 09-10 dataset for the finding of best architectures, then we apply these architecture to the Oli Engineering Statics dataset. In other words, we did not perform architecture search for the Oli Engineering Statics dataset. We believe the performance might be further improved if we also conduct NAS on the Oli statistics dataset. Another reason of taking this approach is we want to see if the architectures found in one dataset could also generalize to another. In other words, we want to see if some modality fusions are helpful when making predictions in unseen datasets. Our implementation code could be found here \footnote{https://github.com/dxywill/multimodal\_nas}.

Our found best model for DKT + FS is shown in Fig. \ref{fig::archs} left. Intuitively, this result can be interpreted as combine \textit{time spent} and \textit{attempts} first using activation function $tanh$, then merge in \textit{first action} using activation function $tanh$. At last, merge in the \textit{response} using activation function $sigmoid$. The best model for NAS extend is shown in Figure \ref{fig::archs} right. This model means using the \textit{first action} as the input for node 1 and use $sigmoid$ activation function. Then, use the \textit{response} as the input for the node 2 and apply activation fucntion $relu$. The output of the node 1 and node 2 will be combined to generate the final output. From the discovered best architectures, both the \textit{response} and the \textit{first action} are selected. We also noticed the performances of these architectures decrease significantly if we exclude the \textit{response} modality. This makes intuitive sense, since a correct response indicates that there is a high probability that this student has mastered the corresponding skill. And the \textit{response} is the only modality used for models like DKT and DKVMN.
The modality \textit{first action} is also selected for two models. This modality indicates if one student chooses to answer the problem or takes other actions, for instance, ask for a hint. If one student asked for a hint, the response will be automatically marked as incorrect. Thus, there are two different situations that may result in an incorrect response and we should be aware the difference. In one situation,  student A  tried to answer the problem but got an incorrect response. On the other hand, student B asked for a hint, thus resulted in an incorrect response. This \textit{first action} modality might reflect the different confidence levels of these two students for some skill. Our model may have learnt assigning different mastery level probabilities for these two situations. Another observation is that using all modalities available might worsen the performance. This is especially true if the dataset itself is noisy. We evaluated the proposed models on two different datasets from the domains of mathematics and engineering separately. However, we believe the proposed models are generalizable to domains as long as skills could be precisely defined.

We conducted 5 fold cross validation and the results are shown in Table \ref{tab::all_results}. We compare our proposed methods with three models using only the correct/incorrect responses (DKT, DKVMN and NAS cell). The NAS cell \cite{ding2020automatic} has a recurrent architecture, but the recurrent unit is automatically designed by neural architecture search. We also compare different ways of fusing different modalities. DKT+SC simply concatenate all the modalities into a big vector. DKT and DKT+SC only differs in the input. DKT+FS as shown in Fig.\ref{fig::search_1} looks for the best fusion architecture. NAS Extend as shown in Fig. \ref{fig::search_2} extend the work \cite{ding2020automatic} to the case of multimodality. As we can see, incorporating more modalities does help improve the performance. All three models using multi modalities (NAS Extend, DKT+FS and DKT+SC) achieved better performance than those only used the response modality. And our proposed model(NAS Extend) that combines the architecture search and model fusion search has the best performance. We also noticed that conducting fusion search (DKT+FS) is not always better than simply concatenation (DKT+SC) as for the Oli Statics 2012 dataset. This could be explained either the fusion search process is not able to fuse these different modalities effectively or the LSTM cell is not able to learn from this fused representation. We performed Mcnemar's test (with $p < 0.01$) on the predictions from the NAS Extend model and the DKT model. The results show significant difference. 

However, we also notice that the improvement is not very big (this is actually one of the reasons we conduct significance tests). One possible explanation is that both datasets are noisy, which makes it hard for the model to learn meaningful relations among modalities. For instance, for the \textit{time spent} feature, there are some transactions with \textit{time spent} values less than 1 second. Thus, we argue better procedures for collecting these data for future research. Besides, it seems the improvement is more obvious on the assistment 09-10 dataset than on the Oli Statistic 2012 Dataset. This could be explained by the fact that we used assistment 09-10 for NAS. 

In short, from the results, the found architectures do generalize to the Oli statics dataset. Besides, we could always run the same methodology separately on a new dataset to see if that could further improve the performance (but we need to consider the tradeoff here, since this process is computational demanding). Because it is possible, due to settings, one modality may have more signal than another in different datasets. In this case, NAS may return different architectures for two different datasets. Readers should be aware of the difference between the generalizability of the models discovered by the methodology and the generalizability of the methodology itself. 

\section{Discussion}

Other than the two proposed approaches of using neural architecture search for incorporating more modalities discussed in the previous sections. We also considered other possibilities. For example, we also tried to extend the LSTM cell for multimodality.  We used one architecture similar to the LSTM cell. However, instead of using the sigmoid and tanh activation functions as in LSTM, we make them tunable ($\phi_1$ and $\phi_2$). We use one such LSTM variant unit for one modality. All the modalities share the same memory cell c and hidden state h. The search process is to find in which order to fuse what modalities and what activation functions to use. Similarly, one sampled architecture could be represented as a list of sequence of numbers. For example, [[$m_1$, $a_1$, $a_0$], [$m_2$, $a_2$, $a_2$]] represents the architecture that first incorporates modality $m_1$ and use activation functions $a_2$ and $a_0$ for $\phi_1$ and $\phi_2$. Then incorporate modality $m_2$ and use activation functions $a_2$ and $a_2$ for $\phi_1$ and $\phi_2$. However, we did not see significant improvements using this methodology, thus their results are not shown in Table \ref{tab::all_results}. More research might be needed to understand if this methodology is useful. 

\section{Conclusion}
In this paper, we discussed ways of using neural architecture search for incorporating multiple modalities for knowledge tracing. We do not aim to develop a new neural architecture search technique for the general purpose, instead we focused on how multiple modalities could fit in neural architecture search, thus improving model prediction performance. We proposed an approach that combines neural architecture search and multimodal fusion within one framework and applied this method for knowledge tracing. We evaluated our proposed approach on two public real datasets, our discovered model is able to achieve superior performance. We did McNemar's test on the predictions and the results are significantly different.  Besides, we argue the importance of measuring how one model performs with time in knowledge tracing and proposed to use time weighted area under the curve (wAUC). We also propose a simple strategy of assigning weights, hoping this will inspire more research work on the evaluation of student response modeling.

\bibliographystyle{ieeetr}
\bibliography{mmas}
\end{document}